# Parsing Free Word-Order Languages in Polynomial Time

Tilman Becker†, Owen Rambow‡

† Institute for Research in Cognitive Science
University of Pennsylvania
tilman@unagi.cis.upenn.edu

‡ TALANA, Université Paris 7
rambow@lifou.linguist.jussieu.fr

## Summary

We present a parsing algorithm with polynomial time complexity for a large subset of V-TAG languages. V-TAG, a variant of multi-component TAG, can handle free-word order phenomena which are beyond the class LCFRS (which includes regular TAG). Our algorithm is based on a CYK-style parser for TAGs.

## 1 Introduction

Long-Distance Scrambling is a word-order phenomenon which is "doubly unbounded" in that (i) more than one element can move, and (ii) movement can be unbounded. In (Becker et al., 1991), we argue that scrambling is beyond TAG by assuming that elementary trees express a complete predicate-argument structure. In (Becker et al., 1992), we show that no formalism in the class LCFRS (which includes TAG) can derive scrambling. (Becker et al., 1991) proposes two variants of the TAG formalism which can derive scrambling while still preserving most of the desirable properties of TAGs (i.e., an extended domain of locality and the factoring of recursion). However, little is known about the formal and computational properties of those systems. (Rambow, 1994) proposes V-TAG, which is closely related to one of the previously proposed varaiants, but redefines the derivation relation.

V-TAG can derive the relevant set of sentences and also cases where scrambling co-occurs with long-distance topicalization (a separate linguistic phenomenon also found in English, in which a single element moves into sentence-initial position):

(1) [Dieses Buch]$_i$ hat [den Kindern]$_j$ bisher noch niemand [PRO t$_j$ t$_i$ zu geben] versucht.
[this book]$_{ACC}$ has [the children]$_{DAT}$ so far yet [no-one]$_{NOM}$ to give tried

So far, no-one has tried to give this book to the children.

We refer to (Müller and Sternefeld, 1993) for a more extensive discussion of the freedom of scrambling in German, Japanese, and Russian. In this paper, we give a parsing algorithm with polynomial time-complexity for lexicalized V-TAG languages.

## 2 V-TAG

Multi-Component TAG (MC-TAG, see (Weir, 1988) for a broader discussion) extends the elementary structures of the grammar from trees to sets[1] of trees. The formal and computational properties of MC-TAG depend on the exact definition of adjunction. "Tree-local" and "set-local" MC-TAG, in which the adjunction sites are restricted, are polynomially parseable, but since they are included in LCFRS, they are not adequate for deriving scrambling (see Section 1). (Weir, 1988) also defines "non-local MC-TAG", in which trees from one set must be adjoined simultaneously anywhere into a derived tree. As shown in (Becker et al., 1991), non-local MC-TAG can handle scrambling. Unfortunately, it is known to generate NP-complete languages (Rambow and Satta, 1992).

---
[1] If a set includes two trees with identical labeling, we assume that the node-addresses are different.

In V-TAG, introduced in (Rambow, 1994), there are no restrictions on adjunction sites. Trees from one tree set can be adjoined anywhere in the derived tree, and they need not be adjoined simultaneously or in a fixed order. Furthermore, trees in the tree sets are equipped with dominance links, first formally defined in (Becker et al., 1991),which have been used previously in linguistic work (for example by (Kroch, 1989)). A dominance link can relate the foot node of a tree to any node in any tree of the same set. The dominance links provide a constraint on possible derivations: after a derivation is completed, each dominance link must hold in the derived tree. Dominance links are essential for encoding structural relations (c-command) between related linguistic elements, such as a head and its arguments.

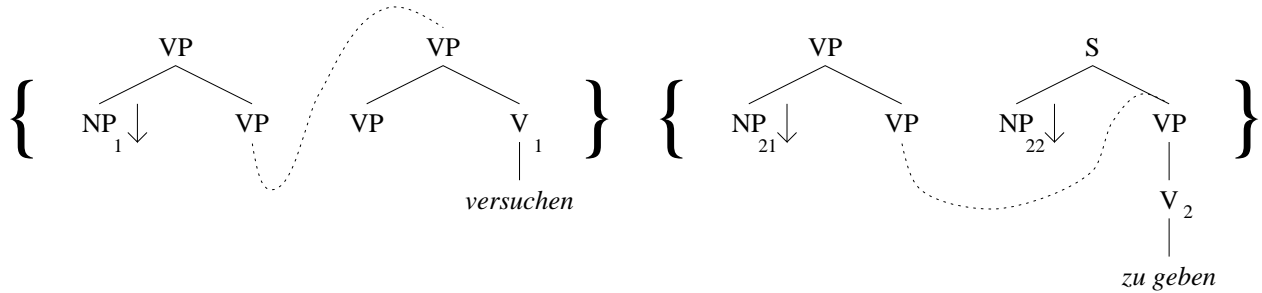

Figure 1: Initial tree set for *versuchen* matrix clause and *geben* embedded clause

For illustrative purposes, we give a V-TAG derivation for sentence (1). The grammar of German is the set of tree sets. Each tree set contains a head (e.g., a verb) and its projections, and slots for its arguments. Two examples are shown in Figure 1. In the set for the *geben* 'to give' embedded clause, one nominal argument is in a separate auxiliary tree, reflecting the fact that it may be scrambled, and the other nominal argument is included in the verbal projection tree, reflecting the fact that it is (long-distance) topicalized. The dotted line represents the dominance link. In the set for the *versuchen* 'to try' matrix clause, the only nominal argument is in a separate auxiliary tree. Its clausal subcategorization requirement is indicated by the fact that the verb is in an auxiliary tree (rooted in VP), forcing adjunction into an embedded clause.

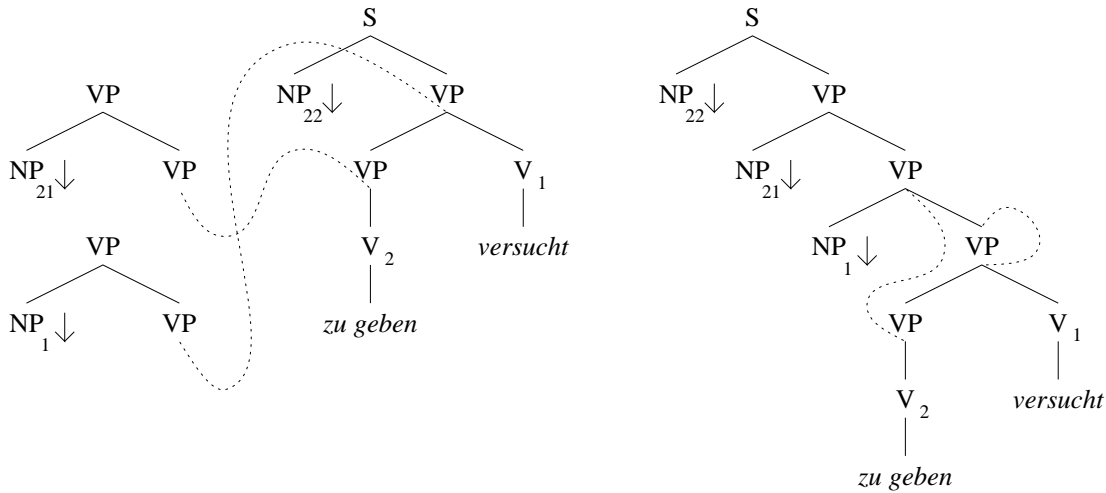

Figure 2: After adjoining matrix clause into subordinate clause (left) and final derived tree (right)

The derivation now proceeds by first adjoining the matrix clause into the embedded clause at the VP node, yielding the structure on the left in Figure 2. This adjunction implements the long-distance topicalization of the embedded direct argument. We are left with two auxiliary trees that still need to be adjoined, representing the scrambled arguments. We first adjoin the matrix subject into its own clause, and then adjoin the embedded indirect object just above the matrix subject. The result is shown in Figure 2 on the right.

Observe that the tree sets given in Figure 1 have the property that they each represent a verb. In linguistic applications of TAG and related formalisms such as V-TAG, it is useful to associate each elementary structure (tree set in the case of V-TAG) with at least one lexical item. Such a grammar is called "lexicalized". This has an important consequence, namely that derivations in a lexicalized grammar are always bounded in length by a linear function of the length of the derived sentence. In the following discussion of a parser for V-TAG, we will make crucial use of this property.

## 3 Parsing V-TAG

In this section, we use an extension of the CYK-type parser for TAG defined by Vijay-Shanker (1987, p.110) to give a polynomial time parser for a large subset of the V-TAG languages. We first describe Vijay-Shanker's parser for simple TAG, and then describe the extensions necessary for V-TAG.

The main idea of Vijay-Shanker's parser is the introduction of a 4-dimensional matrix $T$, in which an entry of a node $\eta$ from an elementary tree $\tau$ at $T[i,j,k,l]$ represents the fact that either

(i) there is some derived tree $\tau'$ such that $\eta$ is its root node and $\eta$ dominates the substring $a_{i+1} \cdots a_j \eta_1 a_k \cdots a_l$ where $\eta_1$ is the (label of the) foot node of $\tau$ or

(ii) there is some derived tree $\tau'$ such that $\eta$ is its root node and $\eta$ dominates the substring $a_{i+1} \cdots a_l$ and $j = k$.

We split every node into a top and a bottom version, similar to the definition of "top" and "bottom" features in a feature-based TAG (Vijay-Shanker, 1987). If $\eta$ is a node in some tree of some set of a VTAG, then $\eta^T$ denotes the top version of that node, and $\eta^B$ the bottom version. The parser fills the matrix $T$ bottom-up, starting from entries for the leaves. (We assume that the grammar is in extended two form, i.e., in every tree every node has at most two children.) There are six cases which fall into two basic categories[2]:
**(i)** Cases 1 to 4 correspond to bottom-up context-free expansions within one elementary tree. Figure 3 shows Case 1.
**(ii)** Cases 5a, 5b, and 6 deal with adjunction. Cases 5a and 5b correspond to adjunction (either at a node which dominates the foot node (5a) or not (5b)). The top version of the node is added to the matrix to reflect the string covered after adjunction at that node has taken place, as illustrated in Figure 3 for Case 5a. Case 6 corresponds to no adjunction: the top version of a node is added if the bottom version is already present in the same cell of the matrix.

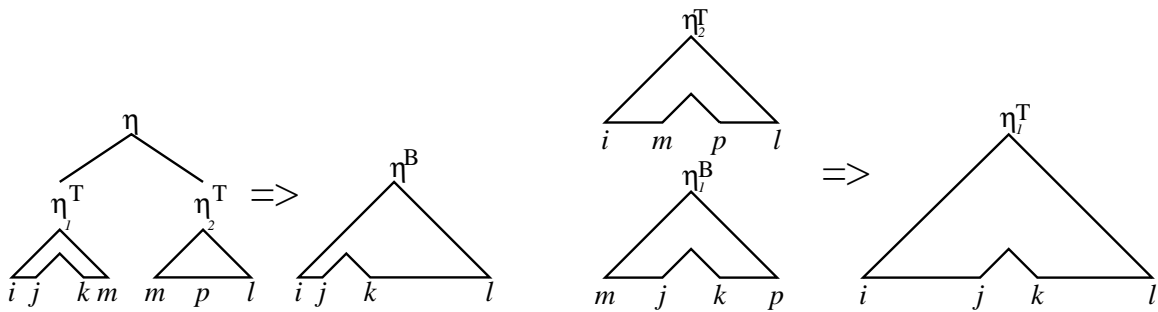

Figure 3: Cases 1 and 5a.

We now turn to the extensions necessary to handle V-TAG. We first introduce some additional terminology. If two nodes $\eta_1$ and $\eta_2$ are linked by a dominance link such that $\eta_1$ dominates $\eta_2$, then we will say that $\eta_1$ has a *passive dominance requirement* and that $\eta_2$ has an *active dominance requirement*. If the tree of which $\eta_1$ ($\eta_2$) is a node has been adjoined during a derivation, but the tree of which $\eta_2$ ($\eta_1$) is a node has not, the dominance requirement (passive or active) will be called *unfulfilled*. The multiset of unfulfilled active dominance requirements of a node $\eta$ will be denoted by $\top(\eta)$, and the multiset of all passive dominance requirements will be denoted by $\bot(\eta)$. We extend this notation to derived trees. Let $\tau$ be a derived tree at any intermediate step of a derivation. We associate with $\tau$ multi-sets which represent all the unfulfilled active and passive dominance requirements of nodes in $\tau$, written $\top(\tau)$ and $\bot(\tau)$, respectively. Observe that a (partial) derived initial tree (i.e., a tree without a footnode on its frontier) cannot have any unfulfilled passive dominance requirements if it is to be part of a successful derivation.

Note that in a lexicalized V-TAG in every derivation $|\top(\tau)|$ and $|\bot(\tau)|$ are always linear with respect to the length of the input string.

In order to keep track of unfulfilled dominance requirements, we add to each entry in the matrix two link-counters which record the number and type of active and passive dominance requirements, respectively, which still need to be satisfied.[3] A link-counter $\gamma$ is an array whose elements are indexed on the dominance links of $G$, and whose values are integers. The sum of two counters is defined component-wise, the norm $|\gamma|$ is defined as the sum of all components. We will denote by $\gamma^\top$ the active requirement counter, by $\gamma^\bot$ the passive requirement counter, and by 0 the counter all of whose values are 0.

---

[2] It is clear how to restrict these cases to implement the adjunction constraints (i.e., obligatory, selective and null adjoining).
[3] This approach is based on a related technique used in (Satta, 1993).

We now spell out what happens to the link-counters in the six cases of the parser. In the following, $a \mathbin{\dot{-}} b$ is defined to be $a - b$ if $a \geq b$, and 0 otherwise.

**Case 1:** $\eta_1$ dominates the foot node (see Figure 3). If there is $(\eta_1^{\mathsf{T}}, \gamma_1^{\perp}, \gamma_1^{\mathsf{T}}) \in T[i, j, k, m]$ and $(\eta_2^{\mathsf{T}}, 0, \gamma_1^{\mathsf{T}}) \in T[m, p, p, l]$, $k \leq m \leq p \leq l$, then add $(\eta^{\mathsf{B}}, \gamma_1^{\perp}, \gamma_1^{\mathsf{T}} + \gamma_2^{\mathsf{T}} + \mathsf{T}(\eta))$ to $T[i, j, k, l]$.

**Cases 2 to 4:** are similar to Case 1.

**Case 5a:** $\eta_1$ dominates the foot node. If there is $(\eta_1^{\mathsf{T}}, \gamma_1^{\perp}, \gamma_1^{\mathsf{T}}) \in T[m, j, k, p]$ and $(\eta_2, \gamma_2^{\perp}, \gamma_2^{\mathsf{T}}) \in T[i, m, p, l]$ where $\eta_2$ is the root node of an auxiliary tree with the same symbol as $\eta_1$, then add $(\eta_1^{\mathsf{T}}, (\gamma_2^{\perp} \mathbin{\dot{-}} \gamma_1^{\mathsf{T}}) + \gamma_1^{\perp}, (\gamma_1^{\mathsf{T}} \mathbin{\dot{-}} \gamma_2^{\perp}) + \gamma_2^{\mathsf{T}})$ to $T[i, j, k, l]$.

**Case 5b:** $\eta_1$ does not dominate the foot node. As 5a, except that then $\gamma_1^{\perp} = 0$, and the move is only valid if $\gamma_2^{\perp} \leq \gamma_1^{\mathsf{T}}$.

**Case 6:** No adjoining takes place at node $\eta$. If there is $(\eta^{\mathsf{B}}, \gamma^{\perp}, \gamma^{\mathsf{T}}) \in T[i, j, k, l]$, then add $(\eta^{\mathsf{T}}, \gamma^{\perp}, \gamma^{\mathsf{T}})$ to $T[i, j, k, l]$.

In all six cases, after calculating the new $\gamma^{\perp}$ and $\gamma^{\mathsf{T}}$, the entry is discarded if $|\gamma^{\perp} + \gamma^{\mathsf{T}}| \geq c \cdot n$, where $c$ is the maximal number of links in a tree set of the grammar. The recognition of a string $a_1 \cdots a_n$ is successful if for some $j$, $0 \leq j \leq n$, and some $\eta$, a root node of an initial tree, we have $(\eta^{\mathsf{T}}, 0, 0) \in T[0, j, j, n]$.

Finally, we can present the algorithm:

Input: $a_1 \cdots a_n, n \geq 0$          Output: ACCEPT/REJECT

---

FOR EVERY $i \in \{0..n-1\}$          "Initialize with leaves"
     IF A LEAF-NODE $\eta$ OF AN ELEMENTARY TREE IS LABELED $a_i$ THEN PUT $(\eta^{\mathsf{T}}, 0, \mathsf{T}(\eta))$ IN $T[i, i+1, i+1, i+1]$
FOR EVERY $i, j \in \{0..n-1\}, i \leq j$          "Initialize with foot nodes"
     FOR EVERY AUXILIARY TREE (WITH FOOT NODE $\eta$): PUT $(\eta^{\mathsf{B}}, \perp(\eta), \mathsf{T}(\eta))$ IN $T[i, i, j, j]$
REPEAT FOR EVERY $i, j, k, l \in \{0..n\}, i \leq j \leq k \leq l$          "parse bottom-up"
     DO CASE $1, 2, 3, 4, 5a, 5b, 6$          "add a new entry"
UNTIL $T$ UNCHANGED
ACCEPT IF $(\eta, 0, 0) \in T[0, j, j, n]$¡, $0 \leq j \leq n$ AND $\eta$ IS ROOT OF SOME INITIAL TREE

**Theorem:** A lexicalized V-TAG is parsable in deterministic polynomial time.

The correctness of the recognition algorithm for TAG is proven by Vijay-Shanker (1987). It can easily be seen by induction on the number of dominance links that the link-counters correctly impose the dominance constraints.

The time complexity of the algorithm is that of Vijay-Shanker's algorithm, $O(n^6)$, multiplied by a factor representing cube of the maximal number of elements of each cell of matrix $T$. Since $|\gamma| \leq c \cdot n$, we have that the number of possible link-counters is bounded by $O(n^{|L|})$ (where $|L|$ is the total number of links in $G$), and the the time complexity of the algorithm is in $O(|G|^3 n^{6|L|} n^6)$, which is polynomial in $n$.

Using back pointers (e.g., for every $(\eta, \gamma)$ which is added to $T$, pointers to the contributing nodes $\eta_1$ and $\eta_2$ in their respective positions are added), the matrix $T$ can be augmented to represent a parse forest from which all derivations of an accepted string can be constructed.